\PassOptionsToPackage{no-math}{fontspec}
\documentclass[displayweek]{sid-arxiv}
\usepackage{array}
\usepackage{stfloats}
\usepackage{microtype}
\usepackage{url}
\hypersetup{pageanchor=false,colorlinks=true,urlcolor=black,citecolor=black,linkcolor=black}
\fnbelowfloat
\makeatletter
\renewenvironment{thebibliography}[1]{%
 \section{References}\small\raggedright
 \list{\@arabic\c@enumiv.}{\settowidth\labelwidth{#1.}%
 \leftmargin\labelwidth\advance\leftmargin\labelsep
 \usecounter{enumiv}\setlength\itemsep{4pt}\setlength\parsep{0pt}}%
 \sloppy\clubpenalty4000\widowpenalty4000\sfcode`\.=1000\relax
}{\endlist}
\makeatother
\begin{document}
\pagenumbering{gobble}
\title{Spatial Memory for AR Glasses: Hierarchical Visual Acquisition and Source-Linked Recall}
\author{Tianxing Zhu\textsuperscript{*}, Dan Huang\textsuperscript{*}\\
\affiliations \textsuperscript{*}Independent Researcher, Shanghai, China}
\date{}
\maketitle
\small

\begin{abstract}
We present a spatial-memory system that enables optical see-through AR glasses to recall earlier physical encounters. The dual-camera prototype combines hierarchical visual acquisition with source-linked storage and staged retrieval. Quality screening, global and regional appearance novelty, and temporal support select monochrome observations for detailed RGB capture. Semantic observations retain links to their source images, allowing retrieval to answer indexed historical questions directly or inspect visual details omitted from scene descriptions. The architecture separates capture acceptance, interpretation outcome, and support for a particular question. This distinction motivates two core control mechanisms: bounded reassessment of familiar views when earlier processing leaves an evidence gap, and retention guided by source dependencies. In a 131.35 s egocentric replay, the implemented acquisition policy selected six RGB views from 648 monitoring observations, while feature extraction ran on 97.7\% of observations. These results demonstrate selective RGB acquisition and identify frequent local inference as a remaining cost. Exposure duration and cross-view timing further constrain the evidence preserved from selected encounters. The system provides a practical acquisition-to-recall architecture for persistent AR assistance, with explicit boundaries between captured evidence, indexed interpretations, and supported historical claims.
\end{abstract}


\section{Introduction}

Persistent memory lets an AI assistant carry information across interactions. Hierarchical context management, editable memory blocks, and durable task artifacts support this continuity in digital environments \cite{r1,r2,r3}. Physical assistance also needs a record of objects, places, and activities encountered outside those interactions. A mug left on a desk or a detail briefly visible on a shelf may never enter a message or document. Answering a later question about that encounter depends on acquiring and retaining its evidence in advance.

Optical see-through AR glasses combine a first-person camera viewpoint with a display available in the physical environment. Our prototype uses RayNeo X3 Pro glasses, with monochrome and RGB cameras and a binocular full-color MicroLED diffractive-waveguide display \cite{r4}. The wearer can retrieve an earlier encounter after the object or scene has left the current view. This requires sufficient visual detail during acquisition and a readable answer within the limited near-eye viewing field.

Acquisition decisions determine what the assistant can remember. Frequent recording accumulates similar views; aggressive selection risks losing brief encounters. Compact scene descriptions make retrieval efficient but can omit details needed by a later question. Interpretation can also fail after an image has been saved. If the capture gate treats that view as adequately remembered, it may suppress a later observation that could fill the gap.

We developed a spatial-memory system that selects monochrome views for detailed RGB capture and stores object-place descriptions with their source images and acquisition times. Retrieval answers from an indexed historical observation when it matches the question, then escalates to image-backed reasoning when further detail is needed. The memory represents physical encounters through these observations and semantic relations.

The contributions are:

1. A hierarchical dual-camera acquisition method that combines global and regional appearance novelty with temporal support to select detailed RGB evidence from monochrome monitoring.

2. A source-linked memory and staged-retrieval architecture that uses searchable interpretations to access original visual evidence.

3. An outcome-aware control design that uses evidence deficits to guide bounded reassessment and source retention.

The prototype implements this complete acquisition-to-recall pipeline. Causal replay, exposure metadata, and memory-processing outcomes characterize how selection and evidence availability depend on camera geometry and acquisition timing.

\section{Related Work}

\subsection{Egocentric episodic memory and historical retrieval:} Ego4D includes episodic-memory queries among its egocentric perception tasks, alongside present-activity understanding and future-event forecasting \cite{r5}. Such queries recover earlier observations of objects or events that may have left the current view. A wearable assistant must preserve the relevant encounter through acquisition and storage decisions made before the question arrives.

Bärmann and Waibel formulate episodic-memory question answering with a constant-size video representation and introduce QAEgo4D \cite{r6}. This makes query-independent encoding and memory capacity explicit. In a wearable system, information can be removed before detailed capture, during semantic interpretation, or through later retention. The resulting losses differ: discarding a caption can leave its source image available, whereas rejecting an encounter before detailed capture removes its fine visual content.

ExoMem combines an AR headset, visual processing, and localization to build an external object-location memory \cite{r7}. ESOM constructs compact egocentric object memory for online visual-query localization \cite{r8}. Both address recall from physical encounters and the construction of memory over time. Our system uses language queries over semantic relations and source images, with acquisition decisions determining the evidence available for recall.

EgoLife studies daily-life assistance using wearable multimodal observations and long-context question answering \cite{r9}. It introduces EgoLifeQA and a retrieval-based assistant architecture. Sustained assistance requires memories that answer questions whose required detail was unknown at acquisition. Our work addresses the capture and evidence management needed for that use.

\subsection{Persistent agent memory and source representations:} MemGPT manages memory tiers to extend effective language-model context \cite{r1}. Letta provides persistent editable memory blocks, and long-running agent harnesses preserve task state in progress artifacts \cite{r2,r3}. Their access and persistence mechanisms support continuity across interactions and inform the organization of wearable memory.

Visual memories also need a record of their origin. A scene description interprets an image captured at a particular time and viewpoint. It indexes that image but may omit readable text, a spatial relation, or a small object requested later. Keeping the source image lets retrieval choose between the semantic index and direct inspection at the required level of detail.

Descriptions, indexes, and derived claims can change while source identity and acquisition time stay fixed. Processing updates therefore preserve the history of physical encounters.

\subsection{Dynamic scene memory and optical see-through AR assistance:} Embodied VideoAgent builds persistent scene memory from egocentric video and embodied sensing, including depth and pose, and updates it with perceived activities \cite{r10}. R4DSG indexes object-centric entries by time, place, persistent identity, and anchor-relative change \cite{r11}. Both represent evolving scenes. Their treatment of identity and temporal change calls for associations across observations as well as descriptions of individual images.

Our memory stores source observations and semantic relations. Class-level sightings support historical recall; identity associations across sources and their observation intervals can support later temporal models.

Memento uses spatiotemporal and activity context to recall remembered interactions and provide proactive AR responses \cite{r12}. Presentation through an optical see-through head-mounted display brings optical constraints. Grubert et al.~survey eye-to-display calibration for aligning virtual content with physical objects \cite{r13}. Gabbard et al.~show how background texture, natural illumination, and drawing style affect outdoor AR text legibility \cite{r14}. Erickson et al.~study interface color schemes under additive display optics, different backgrounds, and lighting conditions \cite{r15}. Gabbard, Mehra, and Swan report performance costs and visual fatigue during repeated context and focal-distance switching \cite{r16}.

Source-linked semantic relations and timestamps allow concise historical recall across presentation formats. World-registered cues can use localization and calibration, while the underlying memory retains the source and time of the encounter.

\subsection{Adaptive acquisition, cascaded sensing, and the capture-memory boundary:} Continuous vision consumes resources in image acquisition, processing, and inference. RedEye moves early convolutional processing into the sensor’s analog domain before quantization, reducing readout and downstream processing costs \cite{r17}. Buckler et al.~study configurable photography and vision modes with reduced sensor precision, resolution, and ISP processing \cite{r18}. Glimpse combines local tracking and scene-change triggers with selective server offloading and cached frames to handle recognition latency \cite{r19}. AdaFrame adapts frame selection for efficient recognition \cite{r20}, while JOCA jointly optimizes camera hardware and adaptive control for downstream tasks \cite{r21}.

We apply staged selection to an open-ended memory task. Low-resolution monochrome monitoring selects detailed RGB acquisitions for later interpretation. Because the question often arrives after the encounter, selection must preserve evidence for future queries while limiting acquisition and interpretation demand.

Visual novelty and memory value can diverge. A novel view may contribute little to a later task, a familiar scene may contain an important small change, and a captured image may fail interpretation. Our outcome-aware controller uses interpretation and retrieval outcomes to reconsider capture decisions when the acquired view has left an evidence gap.

\section{System Architecture and Hierarchical Acquisition}

\subsection{Optical platform and acquisition-to-recall framework:} The prototype runs on RayNeo X3 Pro (ARGF20), whose published display specifications are a binocular diffractive waveguide, full-color MicroLED illumination, 640 \ensuremath{\times} 480 resolution, and a 30° FOV \cite{r4}. Camera metadata reports a 640 \ensuremath{\times} 480 monochrome sensor and a 4032 \ensuremath{\times} 3024 native RGB array. The reference recording uses 640 \ensuremath{\times} 480 monochrome and 1280 \ensuremath{\times} 720 RGB streams. Monochrome monitoring selects encounters, RGB acquisition preserves detail, and the near-eye display presents recalled information.

Acquisition selects detailed images from the available observation history before a future question is known. Interpretation converts these images into source-linked semantic records, and retention determines which source images remain available. When a question arrives, retrieval selects relevant evidence from the retained sources to support an answer.

The architecture distinguishes three states:

\noindent\textbullet\ Acquisition history: visual signatures of views accepted for detailed acquisition.

\noindent\textbullet\ Memory evidence state: retained source images, their processing outcomes, and committed interpretations and quality attributes.

\noindent\textbullet\ Query support: retained evidence sufficient for the claim requested by a particular question.

Separating these states supports asynchronous acquisition, interpretation, and retrieval. An image stays available after interpretation fails, and retrieval can inspect it when a description omits a needed detail.

Camera FOV estimates follow the rectilinear relation $\theta=2\arctan(d/(2f))$, with reported sensor dimension $d$ and focal length $f$ \cite{r22}. These full-array estimates use sensor axes. Stream cropping, lens distortion, and the RGB stream’s 90° presentation rotation affect the displayed camera view; the 30° display FOV belongs to a separate optical path.

Feature computation, detailed acquisition, interpretation requests, transmitted image data, and retained bytes incur separate costs. Selective RGB capture can coexist with frequent local inference.

\subsection{Escalation from monitoring to detailed evidence:} The acquisition hierarchy uses successive tests to decide which observations receive more detailed processing (Figure 1). Quality and visual-change checks operate on the low-resolution monochrome stream. Appearance analysis compares eligible observations with accepted views, and temporal support confirms candidates for RGB acquisition and scene interpretation. Each stage can reject a candidate before further work is incurred.

\begin{table*}[!tb]
\centering
\caption{Optical channels and their roles in the memory system.\label{tab:1}}
\small
\renewcommand{\arraystretch}{1.16}
\setlength{\tabcolsep}{4pt}
\begin{tabular}{|>{\raggedright\arraybackslash}p{\dimexpr 0.18000\linewidth-2\tabcolsep-1.4\arrayrulewidth\relax}|>{\raggedright\arraybackslash}p{\dimexpr 0.44000\linewidth-2\tabcolsep-1.4\arrayrulewidth\relax}|>{\raggedright\arraybackslash}p{\dimexpr 0.38000\linewidth-2\tabcolsep-1.4\arrayrulewidth\relax}|}
\hline
\textbf{Channel} & \textbf{Optical or sensor characteristics} & \textbf{Role and design consequence} \\ \hline
Near-eye display & Manufacturer-specified binocular MicroLED/diffractive waveguide; 640 \ensuremath{\times} 480; 30° FOV \cite{r4} & Concise recall within a limited display field \\ \hline
Monochrome camera & Device-reported 640 \ensuremath{\times} 480; 1.92 \ensuremath{\times} 1.44 mm sensor; 0.71 mm focal length; f/2.0; nominal sensor-axis FOV 107.0° \ensuremath{\times} 90.8° & Broad monitoring view for global and regional selection \\ \hline
RGB camera & Sony IMX681 (manufacturer \cite{r4}); device-reported 4032 \ensuremath{\times} 3024 native array; 4.032 \ensuremath{\times} 3.024 mm sensor; 1.90 mm focal length; f/2.25; nominal sensor-axis FOV 93.4° \ensuremath{\times} 77.0° & Detailed source evidence with a different viewing geometry and exposure state \\ \hline
\end{tabular}
\end{table*}

\begin{figure}[!tb]
\centering
\includegraphics[width=\linewidth,keepaspectratio]{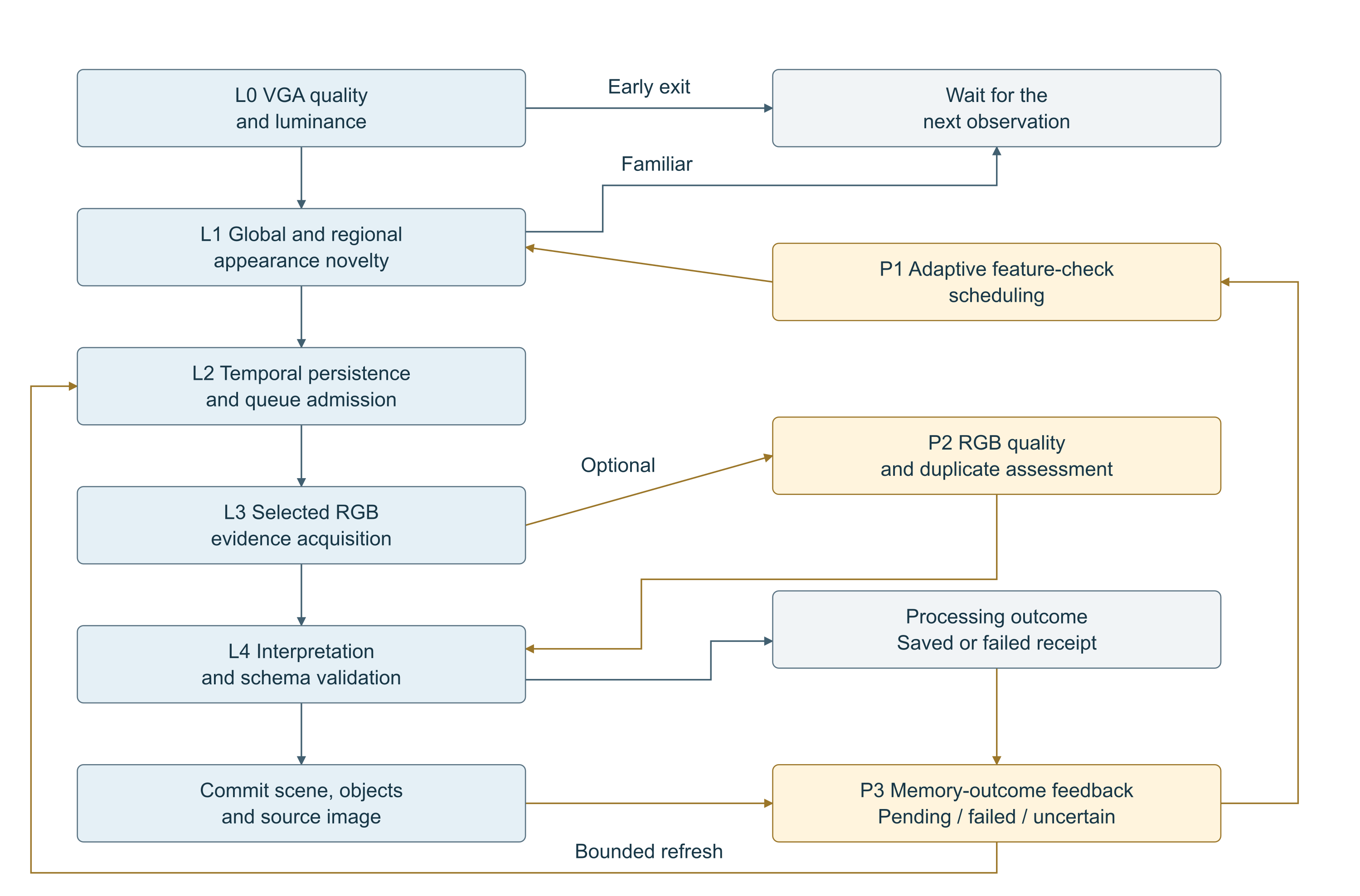}
\caption{Hierarchical capture and evidence feedback.\label{fig:1}}
\end{figure}

The first stage checks brightness and sharpness, then compensates for exposure variation using mean and bounded-gain correction before computing luminance residuals. Whole-view residuals detect broad changes; overlapping patch residuals detect changes confined to smaller regions and can wake feature analysis. Pending candidates continue to receive feature checks after adjacent frames become similar so that temporal support can accumulate.

\subsection{Global and regional novelty:} A mobile image embedder derived from MobileNetV3 and accessed through MediaPipe \cite{r23,r24} produces a normalized full-frame vector $f_t^0$ and four overlapping regional vectors $f_t^r$. For normalized vectors, the distance is $d(\mathbf{u},\mathbf{v})=\operatorname{clip}(1-\mathbf{u}^{\mathsf T}\mathbf{v},0,2).$

For the accepted-view bank $\mathcal B_t$, global novelty is the nearest full-frame distance,

\begin{equation}
G_t=\min_{j\in\mathcal B_t}d(f_t^0,f_j^0).
\end{equation}

Regional comparison uses the $h$ globally nearest reference views, $\mathcal N_h(t)$. Each current region searches their regional vectors. The largest of the resulting nearest-region distances is

\begin{equation}
R_t=\max_{r=1,\ldots,4}\min_{j\in\mathcal N_h(t)}\min_{s=1,\ldots,4}d(f_t^r,f_j^s).
\end{equation}

A candidate satisfies either the global or regional novelty condition:

\begin{equation}
n_t=\mathbf{1}\!\left[G_t\geq\tau_g\;\lor\;(R_t\geq\tau_r\land G_t\geq\tau_{\min})\right].
\end{equation}

The global branch admits broad scene transitions. The regional branch responds to localized differences with some full-view support. Searching across reference regions accommodates appearances that shift between image corners. Taking the largest unmatched regional distance preserves sensitivity to a local difference even when the full view has a close reference.

An empty bank initiates coverage. Accepted views then form a bounded acquisition history. Viewpoint, illumination, and object changes all affect appearance distances; the retained evidence determines their value for later questions.

\subsection{Temporal support and asynchronous acquisition:} A candidate needs a minimum supporting observation count $n_{\min}$ and elapsed duration $T_{\min}$. The count suppresses isolated responses, while duration prevents a dense burst of samples from qualifying as a sustained encounter. An appearance anchor limits feature drift to the threshold \ensuremath{\delta}. Excessive drift starts a new candidate; familiar observations and failed quality checks clear the pending candidate.

Longer temporal support rejects transient triggers but can delay or lose a brief view. In the tested still-capture path, the application closes monochrome monitoring, opens the RGB session, runs a 1 s preview warm-up, acquires a JPEG, and resumes monitoring. Camera closure, session configuration, and exposure timing also contribute to the handoff.

A feature selected in the monochrome view must stay visible and sufficiently resolved in the RGB image. FOV differences, aspect-ratio cropping, camera separation, and head rotation can shift it or move it outside the detailed view. A local pinhole approximation gives

\begin{equation}
\begin{gathered}\Delta u_{\mathrm{rotation}}\simeq f_{\mathrm{px}}|\omega|\Delta t,\\\Delta u_{\mathrm{parallax}}\simeq f_{\mathrm{px}}\frac{|b_\perp|}{Z},\\\ell_{\mathrm{blur}}\simeq f_{\mathrm{px}}|\omega|T_e.\end{gathered}
\end{equation}

where $f_{\mathrm{px}}$ is the local focal scale in pixels, $\omega$ is angular velocity in radians per second, $\Delta t$ is the inter-view acquisition offset, $b_\perp$ is the baseline component transverse to the viewing direction, $Z$ is scene depth, and $T_e$ is exposure time. Rotation and parallax change framing, while exposure integrates motion into blur.

Temporal persistence checks the monochrome candidate before promotion. The RGB quality stage then assesses framing and blur in the detailed source image.

Acquisition waits for downstream processing capacity. Monitoring can resume while interpretation of earlier images is still pending. Candidate persistence follows observation time independently of interpretation completion.

\textbf{Algorithm 1. Causal acquisition with separate capture and semantic commits.}

For each observation x at acquisition time t:

\begin{quote}
\small
if image quality is inadequate: clear candidate; continue

if no wake condition and no candidate needs verification: continue

compute full-frame and regional appearance features

evaluate global novelty G and regional novelty R

if the novelty condition fails: clear candidate; continue

initialize or extend a candidate consistent with its appearance anchor

if temporal support or downstream capacity is insufficient: continue

acquire detailed RGB evidence

if image persistence and queue admission fail: continue

add the triggering signature to acquisition history

interpret the saved image asynchronously

commit a valid source-linked observation or preserve the failure outcome
\end{quote}

Capture acceptance and semantic commitment occur at different times. A saved view can therefore become familiar to the gate before interpretation succeeds.

\section{Source-Linked Memory and Outcome-Aware Coordination}

\subsection{Logical memory hierarchy and invariants:} The memory hierarchy has four roles (Figure 2): working acquisition context (M0), indexed observations (M1), durable source evidence (M2), and derived temporal knowledge (M3).

The observation is the durable unit of evidence, represented as $o_i=(\mathrm{id}_i,t_i,\eta_i,I_i,D_i,E_i,\mu_i),$ where $t_i$ is acquisition time, $\eta_i$ records time quality, $I_i$ is the source image, $D_i$ is the scene description, $E_i$ contains entity-level descriptions and spatial relations, and $\mu_i$ identifies the interpretation model. The source identity remains stable across retrieval and later interpretation.

Scene descriptions provide semantic access to source images. An entity entry records an object class, place, and time; questions about readable text or a disputed relation can return to the image. The semantic index thus provides efficient access while the source preserves further visual detail.

\begin{table*}[!tb]
\centering
\caption{Representational distinctions preserved by the memory hierarchy.\label{tab:2}}
\small
\renewcommand{\arraystretch}{1.16}
\setlength{\tabcolsep}{4pt}
\begin{tabular}{|>{\raggedright\arraybackslash}p{\dimexpr 0.31283\linewidth-2\tabcolsep-1.4\arrayrulewidth\relax}|>{\raggedright\arraybackslash}p{\dimexpr 0.33368\linewidth-2\tabcolsep-1.4\arrayrulewidth\relax}|>{\raggedright\arraybackslash}p{\dimexpr 0.35349\linewidth-2\tabcolsep-1.4\arrayrulewidth\relax}|}
\hline
\textbf{Distinction} & \textbf{Representation} & \textbf{Consequence for recall} \\ \hline
Observation versus interpretation & Stable source identity with a linked description & A revised description remains tied to the original encounter \\ \hline
Acquisition time versus processing time & Capture-time ordering independent of completion & Each response retains the time of its source encounter \\ \hline
Class sighting versus persistent identity & Entity labels and appearance tied to a source & Similar objects can remain identity-ambiguous \\ \hline
Historical versus current state & Time-qualified relations & Reports the observed location and time \\ \hline
Indexed fact versus original detail & Search structure linked to its image & Retrieval can escalate when a question exceeds the description \\ \hline
Processing outcome versus query support & Outcome and quality attributes assessed in query context & A completed record may still be insufficient for a claim \\ \hline
\end{tabular}
\end{table*}

Reprocessing preserves encounter identity, acquisition time orders observations independently of response arrival, and derived claims retain their supporting source interpretations. These update rules preserve encounter history during asynchronous processing.

Memory retains positive observations and uncertain views. A later description that omits an object leaves its earlier sightings intact.

A temporal extension can associate sightings of the same object at different locations at $t_1$ and $t_2$ to represent a change interval $[t_1,t_2]$. The derived record retains those sources and identity assumptions, allowing revision when a supporting source is corrected or removed.

\subsection{Staged evidence access and query support:} Retrieval chooses an access path for the requested claim. Historical location questions first use the index. An eligible class-level sighting returns its recorded place, relation, acquisition time, and source. Ambiguous identity, unsuitable source quality, or a question beyond that indexed relation triggers image inspection.

General questions use a catalog of at most $K$ observations selected for recency and lexical relevance. A selection stage chooses at most $L$ sources for image-backed answering, giving catalog search and visual inspection separate budgets. Descriptions supply temporal and semantic context; images supply details omitted during memory construction.

Retrieval can fail during source selection or image interpretation. A relevant observation may be excluded from the catalog or selected set, or the answer stage may misread a supplied image. Original-image access can recover omitted detail when selection retrieves that source.

Source IDs returned with an answer are checked against the supplied evidence to validate provenance. The sources remain available for inspection. When their content does not support the requested claim, the system can return an insufficient-evidence response.

\begin{figure}[!tb]
\centering
\includegraphics[width=\linewidth,keepaspectratio]{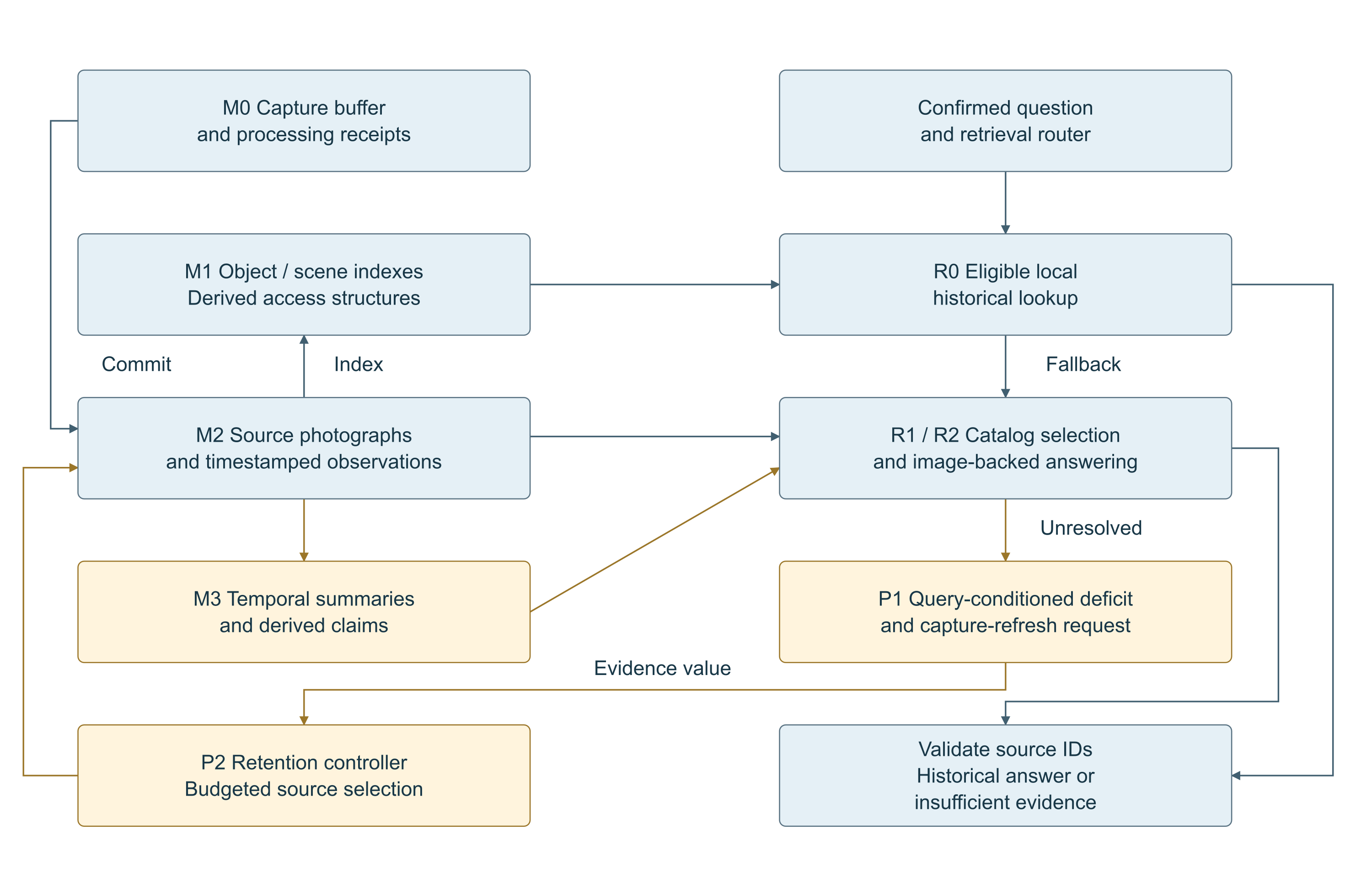}
\caption{Memory organization and staged recall. Indexed observations support local historical lookup; general questions use catalog selection and image-backed answering.\label{fig:2}}
\end{figure}

Support depends on the question. A source may establish “the mug was last seen on the desk” while leaving “the mug is on the desk now” unresolved. A description that lists object classes may omit a serial number, ownership, or exact identity. An image marked globally uncertain may still contain a useful local detail.

Processing outcome records whether a step completed, interpretation quality describes the view, and query support relates a claim to its sources. Keeping these attributes separate allows an imperfect observation to remain useful for questions it can answer.

\begin{table*}[!tb]
\centering
\caption{Outcome-dependent control actions.\label{tab:3}}
\small
\renewcommand{\arraystretch}{1.16}
\setlength{\tabcolsep}{4pt}
\begin{tabular}{|>{\raggedright\arraybackslash}p{\dimexpr 0.31283\linewidth-2\tabcolsep-1.4\arrayrulewidth\relax}|>{\raggedright\arraybackslash}p{\dimexpr 0.33368\linewidth-2\tabcolsep-1.4\arrayrulewidth\relax}|>{\raggedright\arraybackslash}p{\dimexpr 0.35349\linewidth-2\tabcolsep-1.4\arrayrulewidth\relax}|}
\hline
\textbf{Available evidence state} & \textbf{Control action} & \textbf{Reason} \\ \hline
Interpretation pending & Wait and suppress duplicate work & The acquisition has not yet resolved into a semantic outcome \\ \hline
Transient interpretation failure & Consider bounded reuse of the saved image & A suitable saved image can support another interpretation attempt \\ \hline
Visual evidence inadequate and a clearer view recurs & Admit a bounded repeat acquisition & New detail may repair an existing evidence deficit \\ \hline
Committed evidence supports the relevant claim & Preserve the source and suppress redundant work & Additional acquisition has low immediate evidential value \\ \hline
Relevant source excluded by search & Expand or rerank retrieval within its budget & The relevant visual evidence already exists in memory \\ \hline
Past encounter never acquired & Preserve the historical gap & Historical claims require evidence from the relevant encounter \\ \hline
\end{tabular}
\end{table*}

\subsection{Outcome-aware reassessment:} The outcome-aware controller feeds memory outcomes back to acquisition (Figures 1 and 2). It connects each accepted visual signature to its processing outcome and unresolved evidence needs. A familiar view can then be reconsidered when an earlier capture failed to provide adequate memory, while redundant captures remain suppressed.

A control record stores source identity and appearance, processing outcome, evidence attributes, unresolved needs, and the remaining work budget. An outcome becomes eligible for control when processing completes, preserving causal order under variable interpretation delay.

Reassessment chooses between reinterpreting a saved image and acquiring a new one. A suitable image can be reused after processing fails. A new view is useful when it can improve the visual evidence. Retry bounds and termination conditions limit repeated work after systematic failures.

An evidence need records the target, unsupported claim, temporal scope, expiry, and remaining budget. Current-location questions can use fresh observations; historical questions need sources from the relevant interval. Appearance-based place associations identify candidate revisits.

Feature checks can be spaced farther apart during stable, familiar views and brought closer together for a pending candidate or unresolved evidence need. A maximum refresh interval provides checks for local changes missed by the inexpensive wake stage.

\subsection{Source-preserving retention:} Retention determines which sources remain available to future questions. For total memory budget $B_M$, the retained set $S$ must satisfy

\begin{equation}
\begin{gathered}\sum_{i\in S}\operatorname{bytes}(I_i,D_i,E_i)\\+\operatorname{bytes}(\text{access structures and derived records})\\\leq B_M.\end{gathered}
\end{equation}

Source selection weighs unique support, encounter coverage, improved detail, and unresolved needs. Similar-looking images can preserve different text or object states; differently worded descriptions can refer to nearly identical visual evidence.

Retention tracks the evidence affected by each operation. A rebuildable access vector can be discarded, image compression can remove small-object or text detail, and deleting a source can remove the sole support for a claim. Dependent indexes and summaries must then be updated or marked unsupported. Summaries retain selected facts; source images preserve pixel-level detail for later inspection.

\begin{figure*}[!tb]
\centering
\includegraphics[width=0.84\linewidth,keepaspectratio]{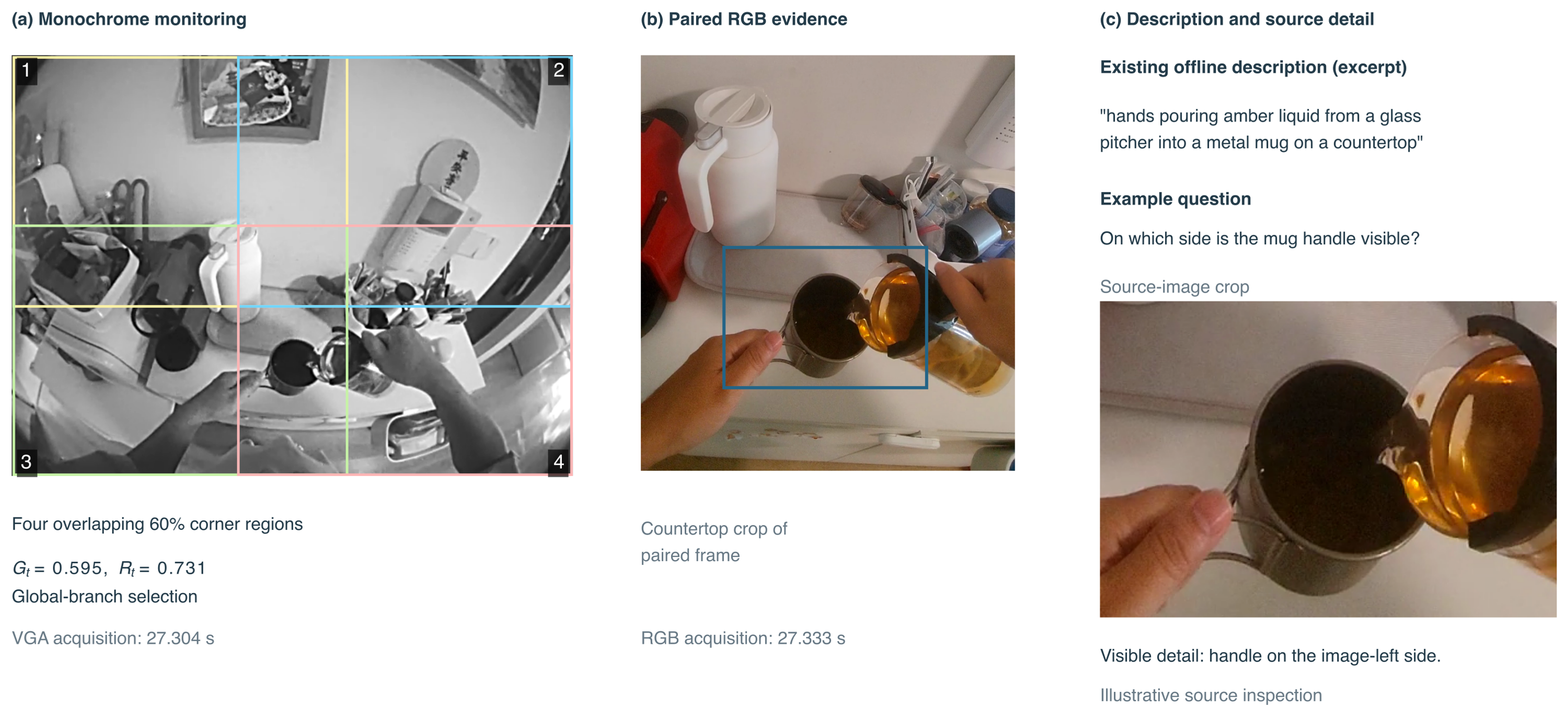}
\caption{Recorded encounter and source-detail walkthrough. (a) The VGA observation with the four overlapping regional encoder crops. Both novelty scores exceed their thresholds, and the global branch admits this candidate. (b) A countertop crop of the paired RGB reference frame, acquired 29.4 ms after the VGA frame and 18.4 ms after the replay request. The cameras provide distinct fields of view. (c) An excerpt from the existing offline audit and an enlarged source detail: the mug handle is visible on the image-left side, a relation absent from the excerpt. The question illustrates source inspection using the recorded reference pair.\label{fig:3}}
\end{figure*}

Evidence dependencies guide source selection. FIFO, uniform sampling, and similarity-based selection provide reference policies at equal retained bytes for assessing their effect on supported recall.

\section{Prototype Characterization and Discussion}

\subsection{Causal replay and selection behavior:} We replayed paired monochrome VGA and RGB reference video from an egocentric walk over a common interval of 131.35 s. VGA observations were processed in source-time order, with each request paired to the first subsequent RGB acquisition within the allowed pairing interval.

In the pouring encounter (Figure 3), the offline description records the activity but omits the mug-handle orientation visible in the source crop. The example question uses that detail to illustrate source inspection.

The policy selected six RGB observations from 648 analyzed VGA observations, or 2.74 selections per minute. Raising the capture ceiling left the count at six, with no budget blocking. The matched-count periodic reference also retained six observations.

Feature extraction ran on 633 observations (97.7\%); the luminance stage bypassed 15 checks (2.3\%). RGB acquisition was selective, but appearance inference still ran on most observations. The adaptive feature scheduler targets this remaining local computation.

The low bypass rate is consistent with continuous head motion during the walk: ego-motion and background parallax shift image content, so pixel-wise luminance residuals can exceed fixed thresholds even when the semantic scene is unchanged. Mean and bounded-gain exposure correction do not remove these geometric changes. Pixel-difference gating is therefore less selective here than in a fixed-camera view. Regional embedding comparison provides an appearance-level test for localized differences, although viewpoint changes also affect embedding distances.

The first RGB observation was accepted at 6.95 s. Initial coverage uses the same persistence rule as later candidates, trading an early capture for stable evidence. The second and sixth acquisitions used the regional branch with global scores below the global threshold (Figure 4). Acquisition also required candidate consistency, temporal support, and queue admission. Scores fell sharply after accepted views entered the acquisition bank.

The continuous reference streams share a sensor clock (Table 4). VGA-to-RGB offsets range from 11.13 to 67.33 ms, with a median of 27.63 ms. Request-to-next-RGB offsets range from 0.48 to 56.90 ms, with a median of 16.80 ms.

\begin{table*}[!tb]
\centering
\caption{Imaging conditions of the six selected RGB reference acquisitions.\label{tab:4}}
\small
\renewcommand{\arraystretch}{1.16}
\setlength{\tabcolsep}{4pt}
\begin{tabular}{|>{\raggedright\arraybackslash}p{\dimexpr 0.20855\linewidth-2\tabcolsep-1.4\arrayrulewidth\relax}|>{\raggedright\arraybackslash}p{\dimexpr 0.26069\linewidth-2\tabcolsep-1.4\arrayrulewidth\relax}|>{\raggedright\arraybackslash}p{\dimexpr 0.26069\linewidth-2\tabcolsep-1.4\arrayrulewidth\relax}|>{\raggedright\arraybackslash}p{\dimexpr 0.27007\linewidth-2\tabcolsep-1.4\arrayrulewidth\relax}|}
\hline
\textbf{RGB time (s)} & \textbf{VGA-to-RGB offset (ms)} & \textbf{Request-to-RGB offset (ms)} & \textbf{RGB exposure (ms)} \\ \hline
6.95 & 11.13 & 0.48 & 10.00 \\ \hline
14.35 & 22.49 & 11.48 & 10.00 \\ \hline
27.33 & 29.39 & 18.41 & 19.99 \\ \hline
56.53 & 67.33 & 56.90 & 19.99 \\ \hline
91.06 & 45.64 & 35.39 & 19.99 \\ \hline
119.40 & 25.88 & 15.19 & 10.00 \\ \hline
\end{tabular}
\end{table*}

At an illustrative head-rotation speed of 60°/s, a 10 to 20 ms exposure integrates 0.6 to 1.2° of angular motion. The median and maximum inter-view offsets correspond to 1.66° and 4.04° of rotation at that speed. A stable monochrome candidate can thus produce a shifted or blurred RGB image, which the post-acquisition quality stage assesses.

\subsection{Interpretation outcomes:} Memory inspection found committed observations marked uncertain and a separate scene-schema failure. The uncertain sources were retained but were ineligible for the local clear-view shortcut. The schema failure retained its image and processing outcome without creating a semantic scene. Saved evidence and indexed memory therefore had distinct processing states.

The prototype uses Xiaomi MiMo-V2.6-Flash for scene descriptions, source selection, and image-backed answers. The glasses call the cloud service over HTTPS; image-quality checks and feature embedding run locally, and source images and memory records remain in app-private storage. No VLM weights are deployed on the glasses. Provider-side precision and quantization are not configured by the client. Eligible historical lookups use the local index without a model call.

The acquisition bank commits when capture is accepted; semantic records depend on later interpretation. The outcome-feedback path uses these processing outcomes to choose between reinterpreting a saved source and reassessing a recurring view.

\begin{figure*}[!tb]
\centering
\includegraphics[width=0.84\linewidth,keepaspectratio]{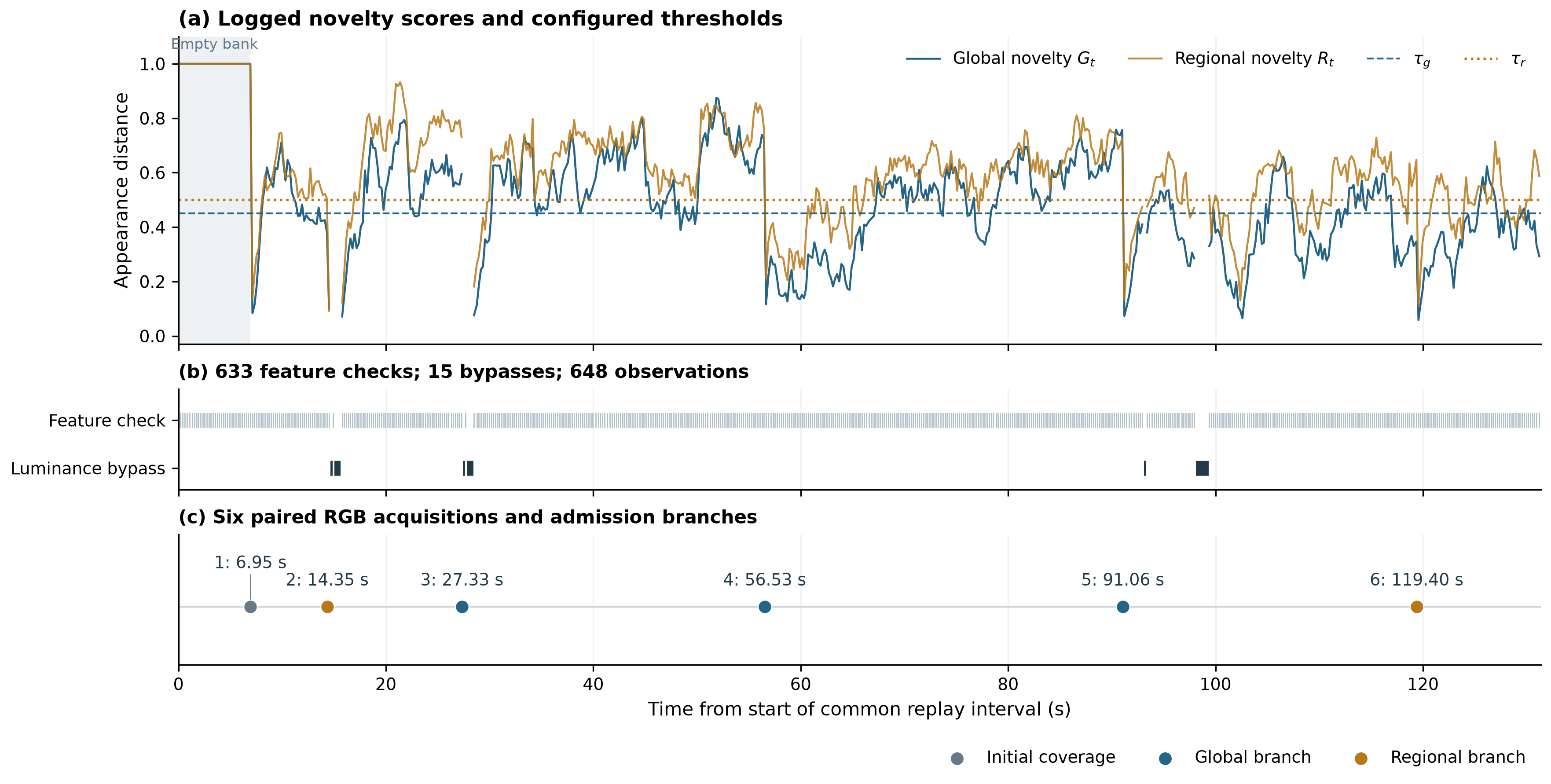}
\caption{Temporal trace of the 131.35 s causal replay. (a) Logged global and regional novelty with the configured thresholds. Gaps mark observations without feature extraction. The shaded initial-coverage interval uses the empty-bank initialization $\boldsymbol{G_t=R_t=1}$. (b) Feature checks and luminance bypasses. (c) Six paired RGB acquisitions, colored by the admitting branch.\label{fig:4}}
\end{figure*}

\subsection{Implications for optical see-through assistance and useful memory:} Recall appears within a 30° near-eye field while the physical scene stays visible. For an additive optical see-through path, the first-order luminance contrast of a bright symbol is

\begin{equation}
C_W\simeq\frac{L_d}{\tau L_b+L_v},
\end{equation}

where $L_d$ is display luminance added at the eye, $L_b$ is scene luminance, $\tau$ is optical transmittance, and $L_v$ is veiling light. A brighter scene raises the background against which symbols must be read. Sparse bright text and low visual density follow from this contrast constraint and the findings on OST legibility and color schemes \cite{r14,r15}. RayNeo advertises 3,500 nit average and 6,000 nit peak brightness \cite{r4}; transmittance and background structure also affect scene-dependent contrast.

Display FOV bounds the visible content, and eye position within the usable eyebox affects visibility. Concise cards place recalled information near the center of the view. World-registered overlays also need eye-to-display calibration and current pose alignment \cite{r13}. A historical object-place relation and timestamp can communicate an earlier encounter directly. The prototype uses concise binocular text; auditory answers could provide a complementary format during visually demanding activity. Real/virtual context changes and focal-distance switching also affect presentation \cite{r16}.

A remembered object-place relation can guide the wearer toward a previously observed location, with its timestamp dating the encounter. Source-linked retrieval and question-specific support preserve this historical meaning in the displayed answer.

Information can be lost when acquisition rejects a brief encounter, interpretation omits a detail, retrieval excludes a source, or retention removes it. Each stage needs a corresponding remedy. Regional analysis addresses localized visual changes, original-image access recovers descriptive omissions, retrieval expansion addresses search misses, and outcome feedback responds to processing or evidence deficits.

Useful memory is assessed through supported recall across the intended questions. Coverage, correctness among answered queries, unsupported present-state assertions, acquisition demand, and retained bytes relate that utility to its acquisition and storage costs.

\section{Conclusion}

We developed an AR-glasses memory system that combines hierarchical visual acquisition, source-linked storage, and staged recall. Capture acceptance, interpretation outcome, and support for a later question have distinct roles in this system. These source-linked paths preserve visual encounters for historical access. Replay shows selective RGB acquisition with frequent local inference; exposure and cross-view timing describe the imaging conditions of the selected evidence. The outcome-aware controller and source-preserving retention policy direct further acquisition and processing toward unresolved evidence needs.

\textbf{Code availability.} The source code for the AR-glasses application is publicly available on GitHub: \href{https://github.com/TZHU64/spatial-memory-ar}{https://github.com/TZHU64/spatial-memory-ar}.

\balance

\end{document}